\documentclass{article}

\PassOptionsToPackage{numbers, compress}{natbib}

\usepackage[preprint]{neurips_2025}

\usepackage[utf8]{inputenc} 
\usepackage[T1]{fontenc}    
\usepackage[hidelinks]{hyperref}       
\usepackage{url}            
\usepackage{booktabs}       
\usepackage{amsfonts}       
\usepackage{nicefrac}       
\usepackage{microtype}      
\usepackage{xcolor}         
\usepackage{colortbl}       
\usepackage{graphicx}
\usepackage[colorinlistoftodos]{todonotes}
\usepackage{array}
\usepackage{xspace}
\usepackage{booktabs}
\usepackage{makecell}
\usepackage{tabularx}
\usepackage{subcaption}

\usepackage[shortcuts]{extdash}   
\renewcommand{\arraystretch}{1.3}

\newcommand{\dataset}{\textsc{VitaGraph}\xspace}
\title{\dataset: Building a Knowledge Graph for Biologically Relevant Learning Tasks}

\author{%
Francesco Madeddu\thanks{Those authors contributed equally.} \\
DIAG,
Sapienza University of Rome\\
IAC, CNR Consiglio Nazionale delle Ricerche\\
  \texttt{madeddu@diag.uniroma1.it} \\
  \And
  Lucia Testa$^\ast$\\
  DIAG,
Sapienza University of Rome\\
   \texttt{lucia.testa@uniroma1.it} \\
  \And
  Gianluca De Carlo$^\ast$\thanks{Work done while the author was on a short-term visit at University of Cambridge, UK.}\\
  DIAG,
Sapienza University of Rome\\
   \texttt{decarlo@diag.uniroma1.it} \\
  \And
  Michele Pieroni\\
  Department of Biochemical Sciences\\
Sapienza University of Rome\\
   \texttt{mi.pieroni@uniroma1.it} \\
   \And
  Andrea Mastropietro\\
 Department of Life Science Informatics and\\Data Science, University of Bonn and
Lamarr Institute\\
   \texttt{mastropietro@bit.uni-bonn.de} \\
   \And
  Aris Anagnostopoulos\\
 DIAG,
Sapienza University of Rome\\
   \texttt{aris.@diag.uniroma1.it} \\
   \And
  Paolo Tieri\\
 IAC,
CNR Consiglio Nazionale delle Ricerche\\
   \texttt{paolo.tieri@cnr.it} \\
   \And
  Sergio Barbarossa\\
 DIET,
Sapienza University of Rome\\
   \texttt{sergio.barbarossa@uniroma1.it} \\
}

\begin{document}

\maketitle

\begin{abstract}
  The intrinsic complexity of human biology presents ongoing challenges to scientific understanding. Researchers worldwide collaborate across disciplines to expand our knowledge of the biological interactions that define human life. Artificial intelligence (AI) methodologies have emerged as powerful tools across scientific domains, particularly in computational biology, where graph data structures effectively model biological entities such as protein--protein interaction (PPI) networks and gene functional networks. Those networks are used as datasets for paramount network medicine tasks, such as gene--disease association prediction, drug repurposing, and polypharmacy side effect studies. Reliable predictions from machine learning models require high-quality foundational data. In this work, we present a comprehensive multi-purpose biological knowledge graph constructed by integrating and refining multiple publicly available datasets. Building upon the Drug Repurposing Knowledge Graph (DRKG), we define an advanced pipeline tasked with a) cleaning inconsistencies and redundancies present in DRKG, b) coalescing information from the main available public data sources, and c) enriching the graph nodes with expressive feature vectors such as molecular fingerprints and gene ontologies, among others. Biologically and chemically relevant features improve the capacity of machine learning models to generate accurate and well-structured embedding spaces. The resulting resource represents a coherent and reliable biological knowledge graph that serves as a state-of-the-art platform to advance research in computational biology and precision medicine. Moreover, it offers the opportunity to benchmark graph-based machine learning and network medicine models on relevant tasks. We demonstrate the effectiveness of the proposed dataset by benchmarking it against the task of drug repurposing, PPI predictions, and side-effect predictions, modeled as link prediction problems. Our dataset, \dataset, is available at: \url{https://www.kaggle.com/datasets/gianlucadecarlods/vitagraph/}.
\end{abstract}

\section{Introduction}
Bioinformatics has undergone remarkable advancements in recent years~\cite{schmidt2024gpus}. This field aims to elucidate complex biological phenomena through the analysis of biological data. Of particular interest are omics data~\cite{perez2017discovering,perez2019quantifying}. The latter encompass molecular information and the interactions among biomolecules across various levels, including genomics, proteomics, and transcriptomics. Notably, the growing interest in understanding protein interactions within organisms has led to the concept of interactome, which involves the construction and analysis of networks representing biological entities and interactions among them. In this context, network medicine~\cite{barabasi2011network} has emerged as a powerful approach for addressing biologically relevant problems by exploiting the structural information encoded in biological networks. One of the most prominent and widely utilized types of such networks is the protein–protein interaction (PPI) network~\cite{oughtred2021biogrid}. In these networks, nodes represent genes or proteins, and edges denote physical interactions. PPI networks, such as BioGRID~\cite{oughtred2021biogrid}, STRING~\cite{szklarczyk2023string}, and HuRI~\cite{luck2020reference}, have been extensively employed, yielding promising results in various applications, including the identification of gene--disease associations (GDAs)~\cite{navlakha2010power, stolfi2023niapu, mastropietro2023xgdag} and the prediction of novel protein--protein interactions~\cite{hu2021survey}. Other relevant types of biological networks include gene regulatory networks~\cite{badia2023gene}, in which nodes representing genes are connected based on their involvement in regulating gene expression, ultimately influencing cellular function. These networks provide complementary information to that offered by PPI networks. Gene–disease networks, typically modeled as bipartite graphs, represent associations between genes and diseases; an edge exists if a gene is implicated in the etiology or pathophysiological mechanisms of a given disease. A relevant example of such a dataset is given by DisGeNET~\citep{pinero2016disgenet,pinero2020disgenet}. Additional relevant types of biomedical networks include drug--drug interaction networks~\cite{knox2024drugbank}, drug--disease networks~\cite{davis2017comparative}, and drug--side effect networks, exemplified by resources such as SIDER~\cite{kuhn2016sider} and OffSides~\cite{tatonetti2012data}. 

Considered individually, these networks---while well-suited to the specific tasks for which they were designed---are insufficient to capture the full complexity of biological systems. Consequently, biological knowledge graphs have been developed with the aim of integrating and harmonizing diverse types of biological information into a unified data resource~\cite{perdomo2024knowledge, callahan2024open} that can better model the complexity of biological organisms. A notable effort in the field of biomedical knowledge graphs is represented by the Drug Repurposing Knowledge Graph (DRKG)~\cite{DRKG}. Originally developed with the aim of identifying potential drug candidates for treating COVID-19~\cite{ioannidis2020few, islam2023molecular}, DRKG integrates a wide range of biomedical data sources. Due to the breadth and diversity of its integrated information, its applicability could be, in principle, extended beyond drug repurposing to a broad range of biologically relevant tasks that can be framed as link prediction problems within the graph learning paradigm.

Despite its potential, the current version of DRKG suffers from numerous inconsistencies that hinder its practical usability and often result in ambiguous or meaningless outcomes. Motivated by the potential offered, we adopted DRKG as the foundation for constructing our proposed knowledge graph. Following a careful and comprehensive cleaning process to address these inconsistencies, we enriched the graph with additional information sourced from reliable and domain-specific biomedical databases. Furthermore, we incorporated biochemically meaningful node features to enhance the expressiveness and biological relevance of the embedded knowledge. We thus propose \dataset (from the Latin word \emph{vita}, meaning \emph{life}), a novel and versatile knowledge graph tailored for a wide range of biological tasks formulated as link prediction problems, including, but not limited to, drug repurposing, gene--disease association, PPI prediction, and side--effect identification. We propose \dataset as a robust benchmark for assessing link prediction methodologies in the context of network medicine. We also provide a customizable pipeline to generate the proposed dataset by including or excluding the steps described in the paper, along with code for benchmark purposes.\footnote[1]{https://github.com/GiDeCarlo/VitaGraph}

\section{Rethinking the Drug Repurposing Knowledge Graph}
Our dataset is built upon DRKG, a comprehensive and heterogeneous biological network that integrates diverse biomedical entities and their interactions. DRKG encompasses a wide array of entity types, including genes, compounds, diseases, biological processes, side effects, symptoms, and other biologically relevant concepts. Its primary aim is to facilitate the exploration of disease mechanisms at the molecular level and to support drug repositioning efforts. The graph aggregates data from six major biomedical databases: DrugBank~\cite{wishart2008drugbank, knox2024drugbank}, Hetionet~\cite{himmelstein2017systematic}, GNBR~\cite{percha2018global}, STRING~\cite{szklarczyk2023string}, IntAct~\cite{del2022intact}, and DGIdb~\cite{griffith2013dgidb, cannon2024dgidb}, as well as curated information from recent literature, including research related to COVID-19. In its latest release, DRKG comprises 97,238 entities spanning 13 distinct types and contains 5,874,261 triplets distributed across 107 edge types. These edge types capture the various relationships that exist among 17 different entity-type pairs, with multiple interaction types possible between the same entity pairs. For example, a compound may interact with a gene both as an inhibitor and as a binder, while gene–gene interactions may include various regulatory and physical associations. This integration of multiple data sources results in a rich, heterogeneous network structure that offers a robust foundation for modeling complex biological systems. As illustrated in Figure~\ref{fig:drkg_before}, DRKG consists of multiple node and edge types, with nodes annotated according to their type and source database, and edges representing specific interaction types between node pairs. 


        

The original DRKG comprises a diverse set of node types, including: anatomy (representing human anatomical structures); ATC (the Anatomical Therapeutic Chemical classification system, used by the World Health Organization to classify drugs); biological process, cellular component, and molecular function (the three branches of the Gene Ontology~\cite{ashburner2000gene}, which describe gene functions); compound (chemical molecules); disease; gene; pathway (a sequence of molecular events within cells that drive cellular functions); pharmacologic class (categorizing drug compounds based on pharmacological properties); side effect; symptom; and tax (taxonomic identifiers indicating the species origin of genes, such as human).  Comprehensive details on the annotation of nodes and edges are provided in Supplementary Material.

Despite its comprehensive scope, the original DRKG exhibits several shortcomings that limit its usability in downstream tasks. These include the presence of duplicate entries, inconsistent formatting standards, heterogeneous or ambiguous labeling, the inclusion of non-human genes, and invalid or outdated compound identifiers. To address these limitations, a rigorous data cleaning phase was necessary prior to the graph’s integration with additional data sources and the enrichment of nodes with biologically and chemically meaningful features.

\subsection{Filtering rows with formatting errors}
 The dataset was analyzed for formatting inconsistencies that could adversely affect downstream processing and analysis. The first issue identified concerned the representation of entity identifiers. While the majority of entries adhered to the format \texttt{entity\_type::database\_source:entity\_id}, a subset lacked the database source. To ensure consistency and facilitate the traceability of entity origins, all identifiers were standardized to follow the former format, explicitly indicating the source database.

Secondly, rows containing the semicolon character (“;”) were flagged, as this character often suggested that multiple entities had been erroneously merged into a single node. These inconsistent entries were removed, resulting in the elimination of 1,122 rows.

Thirdly, the dataset included 98 rows in which compounds were separated by the pipe character (“|”). These cases were interpreted as potential formatting errors or ambiguous representations of compound combinations. Due to the uncertainty surrounding their intended semantics and their limited prevalence, these entries were also excluded from the dataset.

\subsection{Standardization of relationship labels}
As previously noted, the original DRKG comprises 107 distinct edge types (interaction types), many of which are expressed using synonymous terms or alternative codes. To reduce redundancy and enhance semantic consistency across the dataset, we performed a harmonization step in which synonymous interaction labels were mapped to a set of unified and standardized terms. This mapping facilitates more coherent interpretation and analysis of the graph’s relational structure. The complete set of standardized interaction labels and their corresponding original variants is presented in Appendix~\ref{ap:details}.

\subsection{Removal of non-human information}

To focus the dataset on human biology and eliminate biases from COVID-19-related studies, virus-related relationships (e.g., VirGenHumGen, DrugVirGen; see Supplementary Material) were excluded. Non-human genes and their interactions were also removed to avoid introducing noise into human-specific analyses. This filtering step, aimed at enhancing semantic consistency and biological relevance, led to the removal of 135,294 rows and 17,553 non-human genes. For broader research interests, the pipeline allows users to retain non-human interactions by disabling this cleaning step.

\subsection{Standardization of entity identifiers}
The original DRKG was constructed by merging multiple databases that lacked a unified entity identification system, resulting in entities being represented by multiple identifiers. This redundancy led to information loss and fragmentation. To address this issue, we implemented a systematic approach to unify entity identifiers by leveraging supplementary data sources. Mapping entity identifiers poses considerable challenges, particularly because complete cross-database correspondence cannot be ensured, as certain entities may exist exclusively within specific datasets.

\textbf{Compound identifier mapping}
For the standardization of compound identifiers, we employed the UniChem platform~\cite{chambers2013unichem}, which offers comprehensive cross-referencing services by aggregating information from databases such as ChEMBL~\cite{gaulton2012chembl, zdrazil2024chembl} and ChEBI~\cite{degtyarenko2007chebi, hastings2016chebi}. This approach enabled consistent compound mappings across datasets, with PubChem identifiers adopted as the preferred standard due to their extensive coverage~\cite{bolton2008pubchem, li2010pubchem, kim2025pubchem}. Through this standardization process, 2,508 redundant compound IDs were identified and resolved. 

\textbf{Disease mapping}
For disease entity standardization, we leveraged the Human Disease Ontology database~\cite{schriml2019human, schriml2022human}, which provides comprehensive cross-referencing between Disease Ontology (DOID), Medical Subject Headings (MeSH), and Online Mendelian Inheritance in Man (OMIM) identifier systems.Through this standardization process, 118 redundant compound IDs were identified and resolved.

\textbf{Gene mapping}
Genes are identified using the NCBI ID. However, genes retrieved from the DrugBank database may not have a corresponding NCBI ID. To preserve the information carried by such genes, we retained the DrugBank ID for 62 genes for which no mapping was possible.

\subsection{Removal of duplicates}
The steps described above aim to map the largest possible number of entities and interactions to a standardized space, allowing the identification of redundant information. To ensure data consistency, we checked for duplicate edges. In the first stage, exact duplicate triplets are identified and eliminated. However, due to the possibility that certain relationships are recorded with the head and tail nodes in reversed order, an additional step is required. Taking this into account, duplicate entries are detected and removed. The de-duplication procedure resulted in the elimination of 842,262 duplicate rows, thereby significantly improving the structural integrity and consistency of the graph.

\subsection{Additional cleaning procedures}\label{sec:additional_steps}
In the original DRKG, pathway nodes presented several inconsistencies across the various data sources used. Therefore, we initially removed all pathway nodes and their associated connections from the graphs. In Section~\ref{sec:pathways}, we will describe how this information was integrated back into the knowledge graph in a consistent manner. Finally, as an additional cleaning step, nodes representing Taxonomy and Symptoms were removed from the graph, as they were each linked to only a single entity and thus did not contribute additional meaningful information to the knowledge graph. The output of the proposed cleaning pipeline is a refined knowledge graph, free from redundancy, noise, and ambiguous informational content. This intermediate representation serves as a foundational stone for the construction of \dataset, incorporating node features derived and systematically processed from supplementary data sources, as described in the next session.

\section{Enlarging the scope: toward \dataset}\label{sec:vitagraph}
A hallmark of our dataset lies in its integration of additional data sources as well as the definition of biologically meaningful and expressive features. On the one hand, we augment the information content by adding pathway and drug-side effect information from new databases. On the other hand, we enrich the graph nodes with features. Drawing from both biological and chemical domains, the nodes within the graph are endowed with a comprehensive set of descriptors that capture their biochemical significance and role in the biological knowledge. This design paradigm ensures that, beyond the relational structure of the graph, the intrinsic properties of the individual entities contribute substantively to graph learning processes for biological tasks. This enriches the graph structure, leading to \dataset, a new, more comprehensive, and expressive knowledge graph for life science-centered machine learning and network science.

\subsection{Pathways}\label{sec:pathways}
As mentioned in Section~\ref{sec:additional_steps}, pathway nodes were removed due to inconsistencies. To compensate for this, we chose to rely on a single, widely recognized and adopted database: Reactome~\cite{vastrik2007reactome, milacic2024reactome}. Subsequently, we introduced 2,153 new pathway nodes into the graph and connected them to the genes involved in those pathways with 68,380 edges, as detailed in the Reactome database. This resulted in a more consistent and unambiguous integration of molecular pathway knowledge within the graph.

\subsection{Drug--side effect information}
Given the relevance of predicting the side effects of a compound, we integrated the OnSIDES dataset~\cite{tanaka2025onsides} for compound-side effect edges. OnSIDES, curated from real-world pharmacovigilance data, offers a rich source of high-confidence associations that complement existing resources. By incorporating it, we aim to broaden the coverage of known adverse drug reactions, thereby supporting downstream research and analysis with a more comprehensive foundation of pharmacological knowledge. We incorporated the latest version of the dataset, v3.0.0, and selected only the highest-confidence set of edges. To ensure data consistency and avoid introducing noise, we included only those edges involving compounds already present in our graph. Additionally, we mapped and standardized compound and side effect identifiers to prevent redundancy. As a result, a total of 339,867 edges were integrated into the dataset.

\subsection{Elimination of compounds without SMILES representations}
Accurate chemical representation is essential for computational analyses involving molecular compounds. SMILES (Simplified Molecular Input Line Entry System) strings~\cite{smiles} offer a standardized format for describing chemical structures. In this stage, the dataset was cross-referenced with a comprehensive compound dictionary incorporating SMILES annotations curated from DrugBank, GNBR, Hetionet, IntAct, DGIdb, and PubChem. Entries corresponding to compounds lacking an associated SMILES representation were excluded. Such compounds are typically either no longer available (e.g., withdrawn from the market) or proprietary in nature, with undisclosed chemical structures. Consequently, these entries were removed to maintain the openness and reproducibility of research. Moreover, SMILES representations are necessary for subsequent processing steps, such as the generation of Morgan fingerprints (see later), and this led to retaining only those compounds suitable for structural analysis. This filtering process eliminated 3,872 compounds and 239,219 edges.

\subsection{Morgan fingerprint generation} 
To further augment the chemical information content of the knowledge graph, compound nodes were associated with their corresponding Morgan fingerprints~\cite{MorganFinger, RogersFinger}. Morgan fingerprints are a type of circular fingerprint that captures the local structural features of a molecule. By iteratively examining atomic neighborhoods within a defined radius, these fingerprints translate substructural characteristics into fixed-length binary vectors, where each bit denotes the presence (1) or absence (0) of a particular molecular feature. This representation is particularly valuable for similarity searches and machine learning applications in chemoinformatics. Each compound node was assigned a fingerprint feature vector of length 2,048, as is common in the field for machine representation of chemical entities~\cite{capecchi2020one}. We successfully generated and appended 15,831 fingerprints, thereby improving the chemical characterization within the graph.

\subsection{Gene features} 
\begin{figure*}[t]
     \centering
\includegraphics[width = 0.8\linewidth]{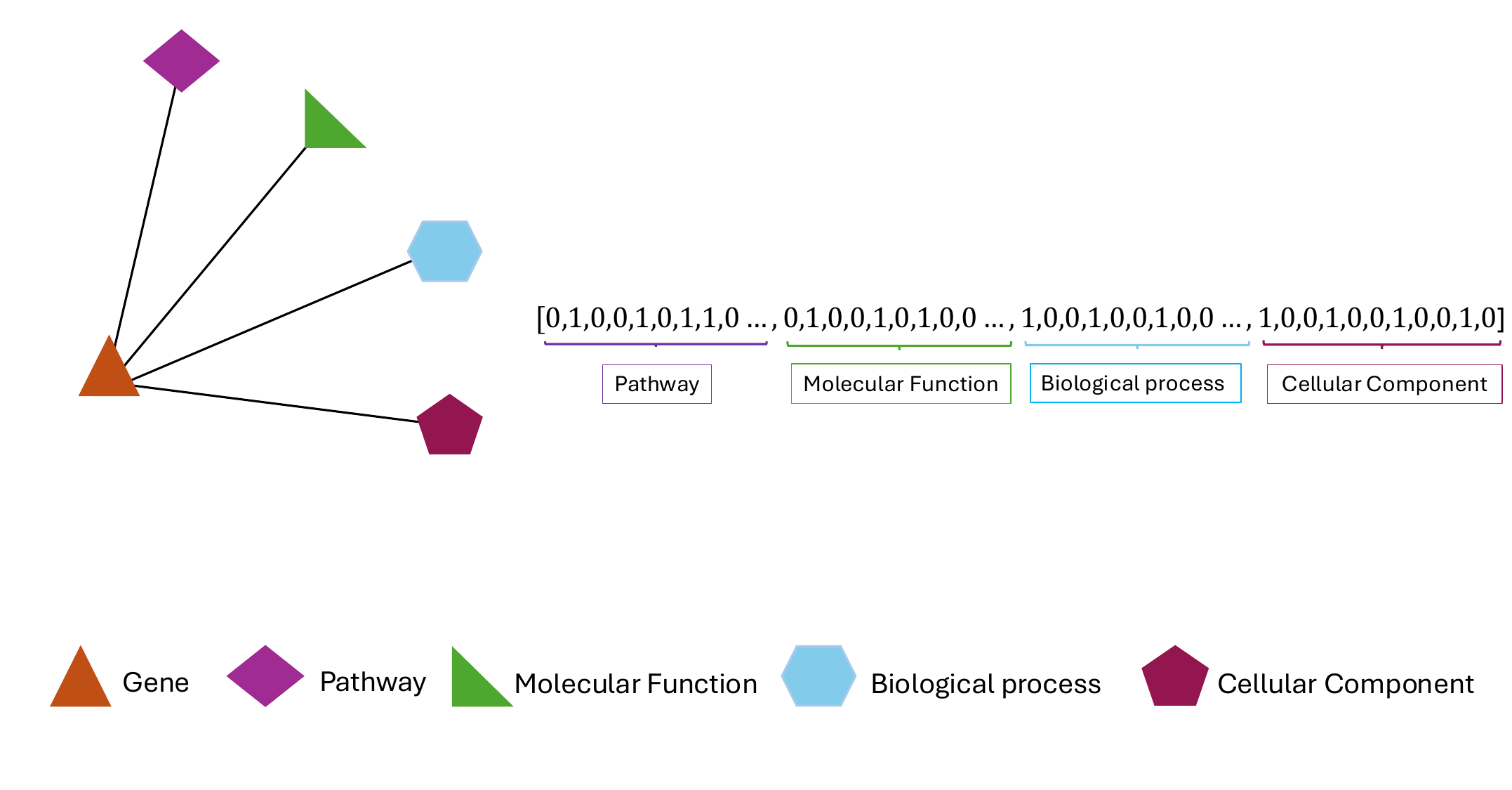}  

        \caption{Construction of the gene feature vector. A one-valued entry in the feature vector means that the gene was connected to a given entity of that type in the original DRKG.}
        
        \label{fig:gene_vector}
\end{figure*}

To incorporate rich biological context, we enhanced the gene nodes by encoding functional information into binary feature vectors. These vectors capture associations with biological pathways, molecular functions, biological processes, and cellular components. In DRKG, such associations were originally represented as edges connecting gene nodes to nodes corresponding to these biological entities. Because nodes representing such entities were linked exclusively to gene nodes and lacked self-loops, thus forming star-shaped subgraphs, we opted to collapse these structures and embed the associated information as gene node features. This approach reduced the graph's complexity while simultaneously enhancing the expressive capacity of the gene nodes. Specifically, 2,153 pathways were encoded using one-hot encoding and incorporated into the gene feature vectors. The same procedure was applied for 2,884 molecular functions, 11,381 biological processes, and 1,391 cellular components. The resulting feature vectors ensure that each gene is accompanied by a detailed functional profile. Figure~\ref{fig:gene_vector} illustrates how such a feature vector is derived from the original graph topology: edges that previously existed in the DRKG are now represented as one-valued entries in the one-hot encoded vector.

\subsection{Final statistics and processing time}
\begin{figure*}[t]
  \centering
  \begin{subfigure}[t]{0.57\textwidth}
    \centering
    \includegraphics[width=\linewidth]{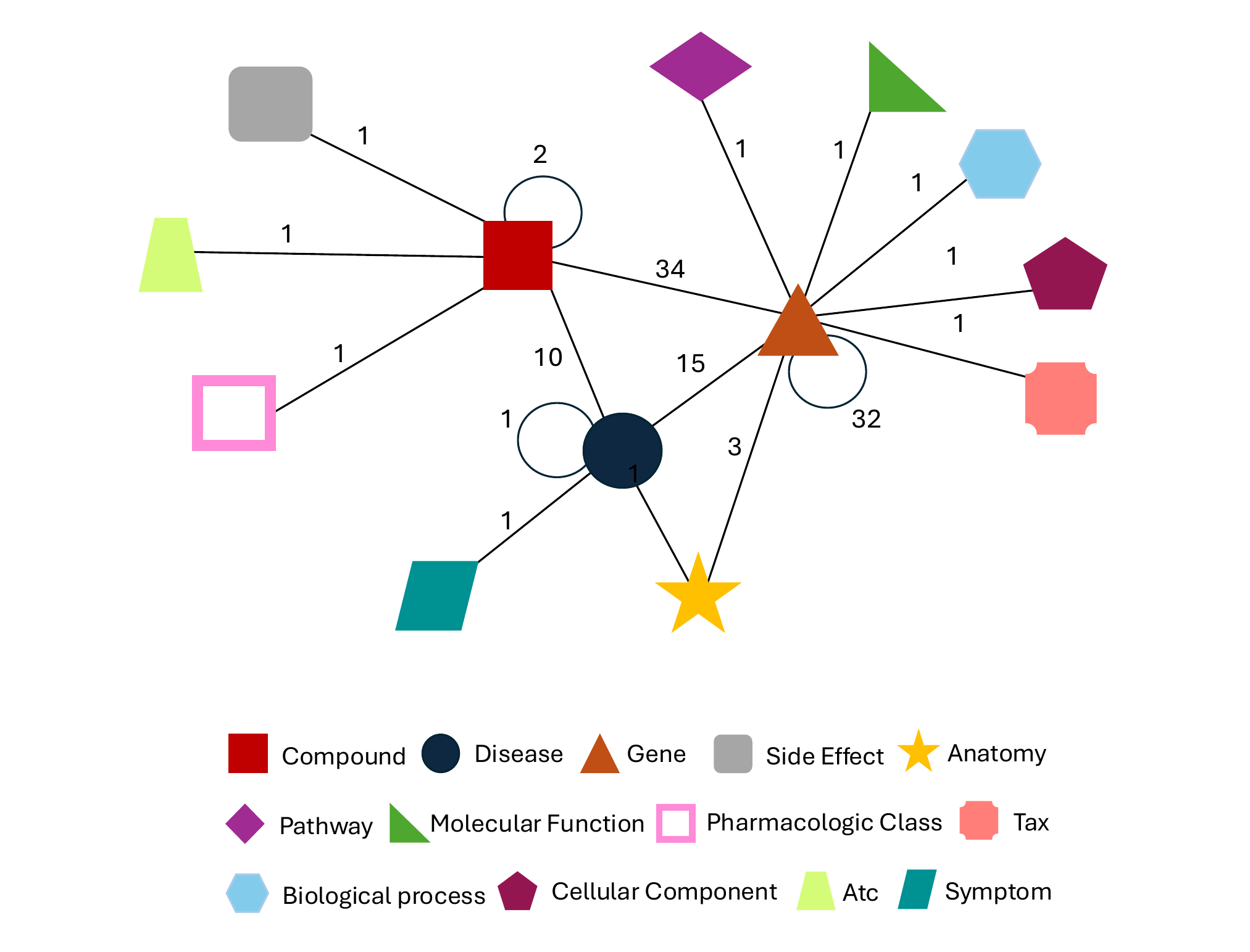}
    \caption{Original DRKG structure.}
    \label{fig:drkg_before}
  \end{subfigure}%
  \hfill
  \begin{subfigure}[t]{0.43\textwidth}
    \centering
    \includegraphics[width=\linewidth]{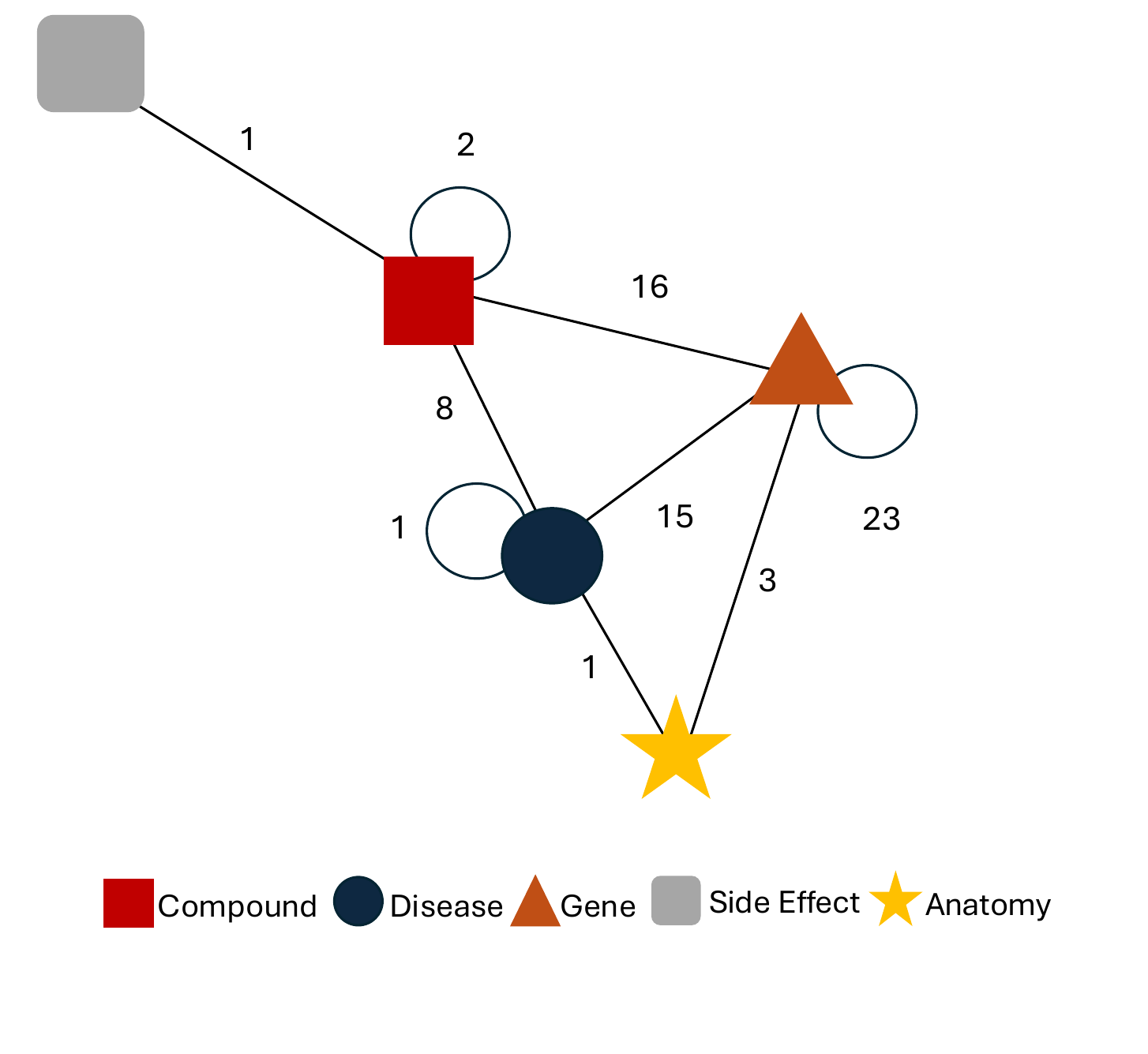}
    \caption{\dataset{} schema.}
    \label{fig:drkg_after}
  \end{subfigure}
  \caption{Comparison of the original DRKG and the new \dataset{} schema. In \dataset schema only a limited number of node and edge types are preserved, augmenting the coherence and reducing the noise within the graph.}
  \label{fig:drkg_comparison}
\end{figure*}

Following cleaning and feature generation procedures, a comparative analysis demonstrated clear and substantial improvements achieved by the newly proposed \dataset. From the original knowledge graph, the total number of rows decreased from 5,874,261 to 4,004,583, corresponding to a reduction of approximately 1,869,678 rows (a 21\% decrease). As a reference, the entire pipeline was executed in 73.27 seconds on an M3 CPU at 4.05~GHz maximum clock, yielding a refined and optimized dataset suitable for advanced computational analyses, machine learning, and data mining for bioinformatics and network medicine tasks. The final graph resulted in 48,058 nodes and 4,004,583 edges. Figure~\ref{fig:drkg_after} shows the final \dataset schema, showcasing its simplified structure.


        
\begin{figure*}[t]
     \centering
\includegraphics[width = 0.8\linewidth]{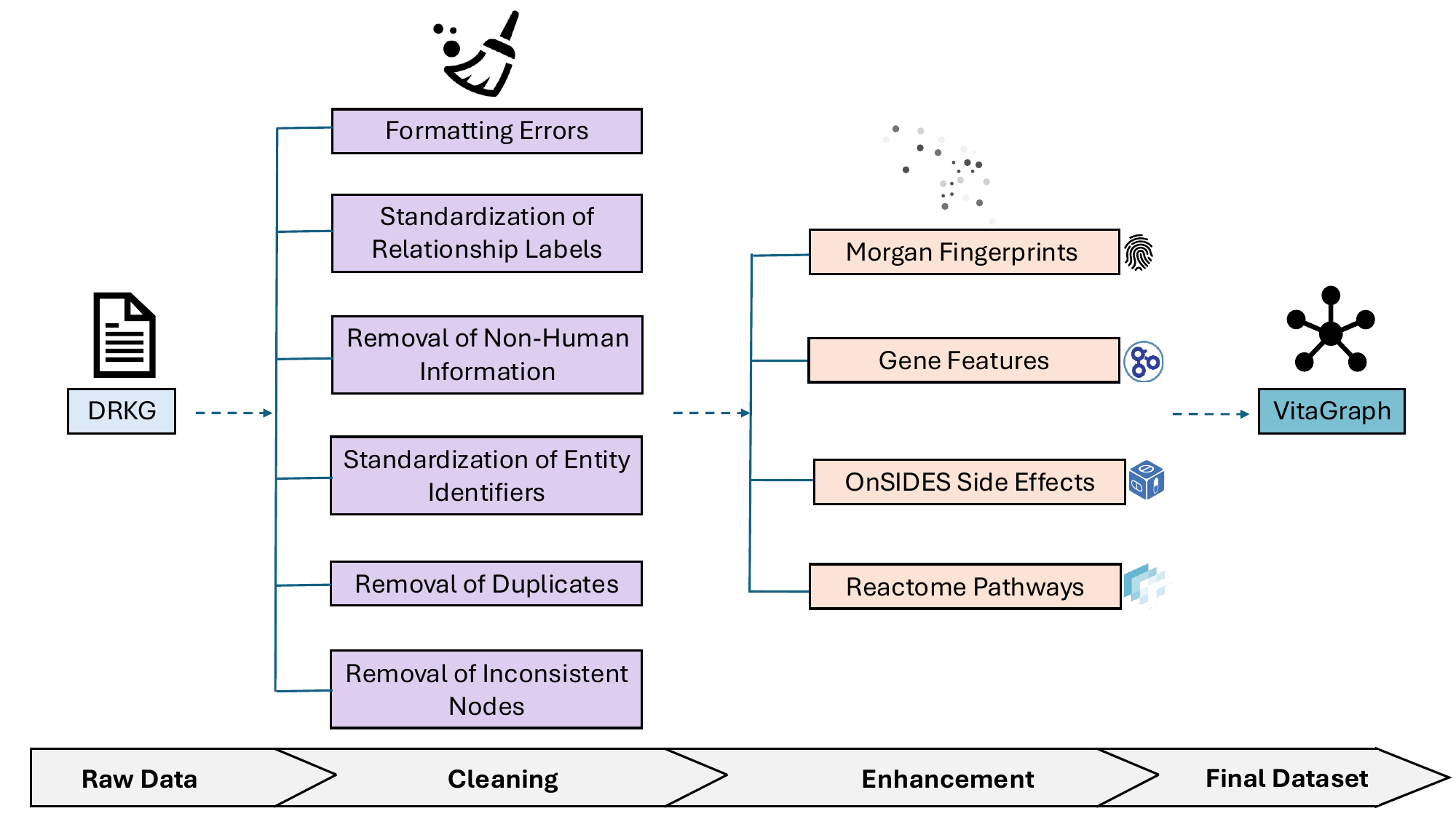}  

        \caption{\dataset creation pipeline.}
        
        \label{fig:pipeline-vitagraph}
\end{figure*}

\section{Benchmarking experiments}
To evaluate the effectiveness of the proposed dataset, we employed three baseline models capable of learning from multi-relational and heterogeneous graph structures. Specifically, we conducted experiments on three prominent link prediction tasks commonly studied in the bioinformatics and network medicine domains, namely drug repurposing, PPI prediction, and drug--side effect detection, modeled as multi-relational link prediction problems, where the objective is not only to determine whether two nodes are connected, but also to predict the type of interaction between them. These tasks serve as strong benchmarks for assessing the utility and generalizability of knowledge graph-based representations in biomedical applications.

To rigorously assess the impact of our dataset enhancements, we performed experiments across three distinct versions of the dataset: (1) the original DRKG dataset, which serves as a baseline, (2) a cleaned version of DRKG including the Reactome and OnSIDES merge without the addition of the features presented in Section~\ref{sec:vitagraph}, and (3) the finally proposed \dataset, which includes both the cleaned structure and node features. This comparative setup allows us to isolate and quantify the contributions of data cleaning and feature enrichment to the predictive performance of the model.

\subsection{Model architectures}\label{sec:models}
As baseline models, we employed modified versions of the relational graph convolutional network (R-GCN)~\cite{schlichtkrull2018modeling} 
and composition-based multi-relational graph convolutional networks (CompGCN)~\cite{vashishth2019composition}, which were adapted to learn from heterogeneous multi-relational graph data. Those architectures are well-suited to the goals of our study, as they can exploit the complex relational structure of biomedical knowledge graphs like ours. Moreover, their flexibility in handling heterogeneous data makes it capable of learning from nodes that possess varying sets of features. To extend the baselines to support heterogeneous nodes, we introduced a node-type-specific linear transformation prior to the graph convolution operations. This preprocessing step projects the feature vectors of different node types into a shared latent space to enable message passing across heterogeneous node types during training. Details on the experimental setup and hyperparameter optimization are provided in Appendix~\ref{ap:setup}. 
\subsection{Training setup}
To train the model on different tasks, specific subsets of triplets were selected. For the PPI task, we used edges connecting pairs of gene entities. For the drug repurposing task, we used edges between compound and gene entities. Finally, for the side effect prediction task, we used edges linking compounds to side effects. While the loss is computed only on these task-specific subsets, all other edges remain in the graph to support message passing and information flow. We adopted a split of 70$\%$ training, 10$\%$ validation, and 20$\%$ test based on the target triplets for each task.

\subsection{Results}

In this section, we present and analyze the results obtained from our experimental setup. Tables~\ref{tab:cg_gc_results}, \ref{tab:cs_results}, \ref{tab:gg_results} report the performance of the models across the various dataset versions and task configurations.

\begin{table}[h!]
  \caption{Drug repurposing results.}
  \label{tab:cg_gc_results}
  \centering
  \scriptsize
  \begin{tabular}{@{} ll cccccc @{}}
    \toprule
    \textbf{Dataset} & \textbf{Model}   & \textbf{AUROC}         & \textbf{AUPRC}        & \textbf{MRR}           & \textbf{Hits@1}         & \textbf{Hits@3}         & \textbf{Hits@10}\\
    \midrule
    DRKG               & R-GCN            & 0.914 ± 0.030          & 0.896 ± 0.028         & 0.148 ± 0.035          & 0.000 ± 0.000          & 0.025 ± 0.075          & 0.650 ± 0.166  \\
                       & CompGCN          & 0.904 ± 0.023          & 0.881 ± 0.028         & 0.246 ± 0.093          & \textbf{0.075 ± 0.115} & 0.350 ± 0.166          & 0.550 ± 0.187  \\
                       \hline
    \dataset (no feat) & R-GCN            & \textbf{0.928 ± 0.004} & \textbf{0.917 ± 0.007}& 0.176 ± 0.090          & 0.000 ± 0.000          & 0.175 ± 0.195          & 0.475 ± 0.305 \\
                       & CompGCN          & 0.827 ± 0.077          & 0.796 ± 0.067         & 0.209 ± 0.121          & 0.050 ± 0.150          & 0.150 ± 0.166          & 0.675 ± 0.225  \\
                       \hline
    \dataset          & R-GCN            & 0.887 ± 0.060          & 0.870 ± 0.053         & 0.219 ± 0.126          & 0.050 ± 0.100          & 0.250 ± 0.354          & 0.825 ± 0.160 \\
                       & CompGCN          & 0.923 ± 0.013          & 0.901 ± 0.016         & \textbf{0.257 ± 0.046} & 0.050 ± 0.100          & \textbf{0.400 ± 0.122} & \textbf{0.925 ± 0.160} \\
    \bottomrule
  \end{tabular}
\end{table}

\begin{table}[h!]
  \caption{Compound--side effect results.}
  \label{tab:cs_results}
  \centering
  \scriptsize
  \begin{tabular}{@{} ll cccccc @{}}
    \toprule
    \textbf{Dataset} & \textbf{Model} & \textbf{AUROC}         & \textbf{AUPRC}        & \textbf{MRR}           & \textbf{Hits@1}         & \textbf{Hits@3}         & \textbf{Hits@10} \\
    \midrule
    DRKG                & R-GCN       & 0.881 ± 0.024          & 0.839 ± 0.034         & 0.126 ± 0.136          & 0.050 ± 0.150          & 0.050 ± 0.150          & 0.300 ± 0.187 \\
                        & CompGCN     & 0.870 ± 0.025          & 0.829 ± 0.025         & 0.158 ± 0.144          & 0.050 ± 0.150          & 0.150 ± 0.200          & 0.300 ± 0.384 \\
    \hline
    \dataset (no feat) & R-GCN       & \textbf{0.899 ± 0.021} & \textbf{0.846 ± 0.024}& \textbf{0.358 ± 0.222}& \textbf{0.250 ± 0.250}& \textbf{0.325 ± 0.251}& \textbf{0.500 ± 0.316} \\
                        & CompGCN     & 0.865 ± 0.027          & 0.807 ± 0.025         & 0.150 ± 0.112          & 0.000 ± 0.000          & 0.200 ± 0.350          & 0.350 ± 0.339 \\
    \hline
    \dataset           & R-GCN       & 0.888 ± 0.019          & 0.827 ± 0.023         & 0.326 ± 0.190          & 0.200 ± 0.245          & \textbf{0.325 ± 0.225}& 0.475 ± 0.261 \\
                        & CompGCN     & 0.824 ± 0.052          & 0.776 ± 0.043         & 0.134 ± 0.069          & 0.000 ± 0.000          & 0.125 ± 0.168          & 0.450 ± 0.245 \\
    \bottomrule
  \end{tabular}
\end{table}

\begin{table}[h!]
  \caption{PPI results.}
  \label{tab:gg_results}
  \centering
  \scriptsize
  \begin{tabular}{@{} ll cccccc @{}}
    \toprule
    \textbf{Dataset} & \textbf{Model} & \textbf{AUROC}         & \textbf{AUPRC}        & \textbf{MRR}           & \textbf{Hits@1}         & \textbf{Hits@3}         & \textbf{Hits@10} \\
    \midrule
    DRKG               & R-GCN       & 0.885 ± 0.032          & 0.871 ± 0.029         & 0.157 ± 0.085          & 0.050 ± 0.100          & 0.100 ± 0.122          & 0.500 ± 0.274       \\
                       & CompGCN     & \textbf{0.924 ± 0.038}& \textbf{0.914 ± 0.041}& 0.245 ± 0.263          & 0.125 ± 0.301          & \textbf{0.275 ± 0.325}& \textbf{0.600 ± 0.300}       \\
                       \hline
    \dataset (no feat) & R-GCN       & 0.893 ± 0.074          & 0.890 ± 0.069         & \textbf{0.280 ± 0.204}& \textbf{0.175 ± 0.225}& \textbf{0.275 ± 0.261}& 0.575 ± 0.354      \\
                        & CompGCN     & 0.859 ± 0.041          & 0.845 ± 0.039         & 0.150 ± 0.055          & 0.000 ± 0.000          & 0.100 ± 0.200          & 0.475 ± 0.261      \\
                        \hline
    \dataset           & R-GCN       & 0.859 ± 0.059          & 0.852 ± 0.054         & 0.139 ± 0.069          & 0.025 ± 0.075          & 0.125 ± 0.168          & 0.475 ± 0.325      \\
                        & CompGCN     & 0.884 ± 0.061          & 0.882 ± 0.059         & 0.180 ± 0.141          & 0.075 ± 0.160          & 0.075 ± 0.160          & 0.475 ± 0.284      \\
    \bottomrule
  \end{tabular}
\end{table}

At first glance, the results on the original DRKG appear promising. However, upon deeper inspection, we demonstrate that these results are significantly affected by data leakage between the training, validation, and test sets.

To identify potential leakage, we examined the three dataset splits by checking for overlap in three ways: 1) duplicate identification, 2) relation-level redundancy, and 3) entity-level redundancy. With the first approach, we searched for identical triplets across splits, including their inverse forms (e.g., if A interacts with B is in the training set, and B interacts with A appears in the test set). This directly indicates leakage of connectivity information. With the second, we apply the pipeline's relation standardization across the splits and check for overlaps. In the original dataset, the same semantic interaction can be represented by multiple redundant relation IDs. Although these IDs are different, they retain the same structural information. As a result, even if a relation appears with a different name in the test set, the model may still recognize and exploit its topology learned from the training set. By standardizing relations, we can identify overlapping interactions that reveal this type of data leakage. Finally, for entity-level redundancy, we standardized entities across splits to detect duplicate nodes. In the original dataset, the same entity may be duplicated under different IDs. These redundant entities typically share similar latent representations and have overlapping neighborhoods. We consider it a form of leakage when an edge is masked for a node in the test set, but the same edge is unmasked for a redundant node in the training set. In this scenario, the model can infer the missing edge based on the connections of a nearly identical redundant entity's latent representation. Table~\ref{tab:leakage} quantifies the extent of data leakage across tasks. We find substantial leakage in the PPI task and the drug repurposing task, significantly compromising the reliability of the results. In contrast, the side effect prediction task remains unaffected. This is because its edges---which connect compounds to side effects---all originate from the same source dataset (SIDER) and are not subject to redundancy.




\begin{table}[h!]
  \caption{Leakage interaction ratios for train/val and train/test splits.  No leakage is present for the compound--side effect task.}
  \label{tab:leakage}
  \centering
  \scriptsize                    
  \setlength{\tabcolsep}{10pt}
  \renewcommand{\arraystretch}{1.5}
  \begin{tabular}{@{}lccc@{}}
    \toprule
    \textbf{Split}   & \textbf{PPI} & \textbf{Drug repurposing} & \textbf{Compound-side effect}\\
    \midrule
    Train Val        & 0.655 $\pm$ 0.002        & 0.182 $\pm$ 0.003   &   --            \\
    Train Test       & 0.655  $\pm$ 0.001      & 0.187 $\pm$ 0.002    &   --        \\
    \bottomrule
  \end{tabular}
\end{table}

The results show that our approach achieves performance comparable to, or better than, the original dataset. However, a fair comparison is not possible due to the substantial data leakage present in the original dataset. Constructing a leakage-free version of that dataset would result in a structure very similar to our proposed version, albeit without the additional enrichment we introduce.

\section{Conclusions}
In this work, we introduce \dataset, a knowledge graph specifically designed to support graph machine learning in biologically and chemically relevant applications. Building upon the DRKG, we perform comprehensive data cleaning and node enrichment by incorporating biochemically meaningful features derived from diverse data sources. The resulting knowledge graph is intended to serve the research community in the discovery of novel biological insights, such as previously unidentified gene--gene interactions or drug repurposing opportunities, treated as link prediction problems. Our results demonstrate that relational graph neural network models can effectively learn from \dataset, indicating its suitability as a benchmark for graph machine learning and knowledge extraction models. Moreover, novel connections discovered through this framework, once experimentally validated, can be reintegrated into the graph, thereby continually enhancing the breadth and depth of biomedical knowledge it contains. One limitation of our dataset lies in its reliance on the quality of the integrated data sources. Nevertheless, by maintaining regular updates, we aim to provide researchers with a reliable and continually improving resource for biomedical discovery.



\clearpage
\bibliographystyle{unsrtnat}
\bibliography{neurips_2025}

\begin{thebibliography}{49}
\providecommand{\natexlab}[1]{#1}
\providecommand{\url}[1]{\texttt{#1}}
\expandafter\ifx\csname urlstyle\endcsname\relax
  \providecommand{\doi}[1]{doi: #1}\else
  \providecommand{\doi}{doi: \begingroup \urlstyle{rm}\Url}\fi

\bibitem[Schmidt and Hildebrandt(2024)]{schmidt2024gpus}
Bertil Schmidt and Andreas Hildebrandt.
\newblock From gpus to ai and quantum: three waves of acceleration in bioinformatics.
\newblock \emph{Drug Discovery Today}, page 103990, 2024.

\bibitem[Perez-Riverol et~al.(2017)Perez-Riverol, Bai, da~Veiga~Leprevost, Squizzato, Park, Haug, Carroll, Spalding, Paschall, Wang, et~al.]{perez2017discovering}
Yasset Perez-Riverol, Mingze Bai, Felipe da~Veiga~Leprevost, Silvano Squizzato, Young~Mi Park, Kenneth Haug, Adam~J Carroll, Dylan Spalding, Justin Paschall, Mingxun Wang, et~al.
\newblock Discovering and linking public omics data sets using the omics discovery index.
\newblock \emph{Nature biotechnology}, 35\penalty0 (5):\penalty0 406--409, 2017.

\bibitem[Perez-Riverol et~al.(2019)Perez-Riverol, Zorin, Dass, Vu, Xu, Glont, Vizca{\'\i}no, Jarnuczak, Petryszak, Ping, et~al.]{perez2019quantifying}
Yasset Perez-Riverol, Andrey Zorin, Gaurhari Dass, Manh-Tu Vu, Pan Xu, Mihai Glont, Juan~Antonio Vizca{\'\i}no, Andrew~F Jarnuczak, Robert Petryszak, Peipei Ping, et~al.
\newblock Quantifying the impact of public omics data.
\newblock \emph{Nature communications}, 10\penalty0 (1):\penalty0 3512, 2019.

\bibitem[Barab{\'a}si et~al.(2011)Barab{\'a}si, Gulbahce, and Loscalzo]{barabasi2011network}
Albert-L{\'a}szl{\'o} Barab{\'a}si, Natali Gulbahce, and Joseph Loscalzo.
\newblock Network medicine: a network-based approach to human disease.
\newblock \emph{Nature reviews genetics}, 12\penalty0 (1):\penalty0 56--68, 2011.

\bibitem[Oughtred et~al.(2021)Oughtred, Rust, Chang, Breitkreutz, Stark, Willems, Boucher, Leung, Kolas, Zhang, et~al.]{oughtred2021biogrid}
Rose Oughtred, Jennifer Rust, Christie Chang, Bobby-Joe Breitkreutz, Chris Stark, Andrew Willems, Lorrie Boucher, Genie Leung, Nadine Kolas, Frederick Zhang, et~al.
\newblock The biogrid database: A comprehensive biomedical resource of curated protein, genetic, and chemical interactions.
\newblock \emph{Protein Science}, 30\penalty0 (1):\penalty0 187--200, 2021.

\bibitem[Szklarczyk et~al.(2023)Szklarczyk, Kirsch, Koutrouli, Nastou, Mehryary, Hachilif, Gable, Fang, Doncheva, Pyysalo, et~al.]{szklarczyk2023string}
Damian Szklarczyk, Rebecca Kirsch, Mikaela Koutrouli, Katerina Nastou, Farrokh Mehryary, Radja Hachilif, Annika~L Gable, Tao Fang, Nadezhda~T Doncheva, Sampo Pyysalo, et~al.
\newblock The string database in 2023: protein--protein association networks and functional enrichment analyses for any sequenced genome of interest.
\newblock \emph{Nucleic acids research}, 51\penalty0 (D1):\penalty0 D638--D646, 2023.

\bibitem[Luck et~al.(2020)Luck, Kim, Lambourne, Spirohn, Begg, Bian, Brignall, Cafarelli, Campos-Laborie, Charloteaux, et~al.]{luck2020reference}
Katja Luck, Dae-Kyum Kim, Luke Lambourne, Kerstin Spirohn, Bridget~E Begg, Wenting Bian, Ruth Brignall, Tiziana Cafarelli, Francisco~J Campos-Laborie, Benoit Charloteaux, et~al.
\newblock A reference map of the human binary protein interactome.
\newblock \emph{Nature}, 580\penalty0 (7803):\penalty0 402--408, 2020.

\bibitem[Navlakha and Kingsford(2010)]{navlakha2010power}
Saket Navlakha and Carl Kingsford.
\newblock The power of protein interaction networks for associating genes with diseases.
\newblock \emph{Bioinformatics}, 26\penalty0 (8):\penalty0 1057--1063, 2010.

\bibitem[Stolfi et~al.(2023)Stolfi, Mastropietro, Pasculli, Tieri, and Vergni]{stolfi2023niapu}
Paola Stolfi, Andrea Mastropietro, Giuseppe Pasculli, Paolo Tieri, and Davide Vergni.
\newblock Niapu: network-informed adaptive positive-unlabeled learning for disease gene identification.
\newblock \emph{Bioinformatics}, 39\penalty0 (2):\penalty0 btac848, 2023.

\bibitem[Mastropietro et~al.(2023)Mastropietro, De~Carlo, and Anagnostopoulos]{mastropietro2023xgdag}
Andrea Mastropietro, Gianluca De~Carlo, and Aris Anagnostopoulos.
\newblock Xgdag: explainable gene--disease associations via graph neural networks.
\newblock \emph{Bioinformatics}, 39\penalty0 (8):\penalty0 btad482, 2023.

\bibitem[Hu et~al.(2021)Hu, Wang, Huang, Hu, and You]{hu2021survey}
Lun Hu, Xiaojuan Wang, Yu-An Huang, Pengwei Hu, and Zhu-Hong You.
\newblock A survey on computational models for predicting protein--protein interactions.
\newblock \emph{Briefings in bioinformatics}, 22\penalty0 (5):\penalty0 bbab036, 2021.

\bibitem[Badia-i Mompel et~al.(2023)Badia-i Mompel, Wessels, M{\"u}ller-Dott, Trimbour, Ramirez~Flores, Argelaguet, and Saez-Rodriguez]{badia2023gene}
Pau Badia-i Mompel, Lorna Wessels, Sophia M{\"u}ller-Dott, R{\'e}mi Trimbour, Ricardo~O Ramirez~Flores, Ricard Argelaguet, and Julio Saez-Rodriguez.
\newblock Gene regulatory network inference in the era of single-cell multi-omics.
\newblock \emph{Nature Reviews Genetics}, 24\penalty0 (11):\penalty0 739--754, 2023.

\bibitem[Pi{\~n}ero et~al.(2016)Pi{\~n}ero, Bravo, Queralt-Rosinach, Guti{\'e}rrez-Sacrist{\'a}n, Deu-Pons, Centeno, Garc{\'\i}a-Garc{\'\i}a, Sanz, and Furlong]{pinero2016disgenet}
Janet Pi{\~n}ero, {\`A}lex Bravo, N{\'u}ria Queralt-Rosinach, Alba Guti{\'e}rrez-Sacrist{\'a}n, Jordi Deu-Pons, Emilio Centeno, Javier Garc{\'\i}a-Garc{\'\i}a, Ferran Sanz, and Laura~I Furlong.
\newblock Disgenet: a comprehensive platform integrating information on human disease-associated genes and variants.
\newblock \emph{Nucleic acids research}, page gkw943, 2016.

\bibitem[Pi{\~n}ero et~al.(2020)Pi{\~n}ero, Ram{\'\i}rez-Anguita, Sa{\"u}ch-Pitarch, Ronzano, Centeno, Sanz, and Furlong]{pinero2020disgenet}
Janet Pi{\~n}ero, Juan~Manuel Ram{\'\i}rez-Anguita, Josep Sa{\"u}ch-Pitarch, Francesco Ronzano, Emilio Centeno, Ferran Sanz, and Laura~I Furlong.
\newblock The disgenet knowledge platform for disease genomics: 2019 update.
\newblock \emph{Nucleic acids research}, 48\penalty0 (D1):\penalty0 D845--D855, 2020.

\bibitem[Knox et~al.(2024)Knox, Wilson, Klinger, Franklin, Oler, Wilson, Pon, Cox, Chin, Strawbridge, et~al.]{knox2024drugbank}
Craig Knox, Mike Wilson, Christen~M Klinger, Mark Franklin, Eponine Oler, Alex Wilson, Allison Pon, Jordan Cox, Na~Eun Chin, Seth~A Strawbridge, et~al.
\newblock Drugbank 6.0: the drugbank knowledgebase for 2024.
\newblock \emph{Nucleic acids research}, 52\penalty0 (D1):\penalty0 D1265--D1275, 2024.

\bibitem[Davis et~al.(2017)Davis, Grondin, Johnson, Sciaky, King, McMorran, Wiegers, Wiegers, and Mattingly]{davis2017comparative}
Allan~Peter Davis, Cynthia~J Grondin, Robin~J Johnson, Daniela Sciaky, Benjamin~L King, Roy McMorran, Jolene Wiegers, Thomas~C Wiegers, and Carolyn~J Mattingly.
\newblock The comparative toxicogenomics database: update 2017.
\newblock \emph{Nucleic acids research}, 45\penalty0 (D1):\penalty0 D972--D978, 2017.

\bibitem[Kuhn et~al.(2016)Kuhn, Letunic, Jensen, and Bork]{kuhn2016sider}
Michael Kuhn, Ivica Letunic, Lars~Juhl Jensen, and Peer Bork.
\newblock The sider database of drugs and side effects.
\newblock \emph{Nucleic acids research}, 44\penalty0 (D1):\penalty0 D1075--D1079, 2016.

\bibitem[Tatonetti et~al.(2012)Tatonetti, Ye, Daneshjou, and Altman]{tatonetti2012data}
Nicholas~P Tatonetti, Patrick~P Ye, Roxana Daneshjou, and Russ~B Altman.
\newblock Data-driven prediction of drug effects and interactions.
\newblock \emph{Science translational medicine}, 4\penalty0 (125):\penalty0 125ra31--125ra31, 2012.

\bibitem[Perdomo-Quinteiro and Belmonte-Hern{\'a}ndez(2024)]{perdomo2024knowledge}
Pablo Perdomo-Quinteiro and Alberto Belmonte-Hern{\'a}ndez.
\newblock Knowledge graphs for drug repurposing: a review of databases and methods.
\newblock \emph{Briefings in Bioinformatics}, 25\penalty0 (6):\penalty0 bbae461, 2024.

\bibitem[Callahan et~al.(2024)Callahan, Tripodi, Stefanski, Cappelletti, Taneja, Wyrwa, Casiraghi, Matentzoglu, Reese, Silverstein, et~al.]{callahan2024open}
Tiffany~J Callahan, Ignacio~J Tripodi, Adrianne~L Stefanski, Luca Cappelletti, Sanya~B Taneja, Jordan~M Wyrwa, Elena Casiraghi, Nicolas~A Matentzoglu, Justin Reese, Jonathan~C Silverstein, et~al.
\newblock An open source knowledge graph ecosystem for the life sciences.
\newblock \emph{Scientific Data}, 11\penalty0 (1):\penalty0 363, 2024.

\bibitem[Ioannidis et~al.(2020{\natexlab{a}})Ioannidis, Song, Manchanda, Li, Pan, Zheng, Ning, Zeng, and Karypis]{DRKG}
Vassilis~N. Ioannidis, Xiang Song, Saurav Manchanda, Mufei Li, Xiaoqin Pan, Da~Zheng, Xia Ning, Xiangxiang Zeng, and George Karypis.
\newblock Drkg - drug repurposing knowledge graph for covid-19.
\newblock \url{https://github.com/gnn4dr/DRKG/}, 2020{\natexlab{a}}.

\bibitem[Ioannidis et~al.(2020{\natexlab{b}})Ioannidis, Zheng, and Karypis]{ioannidis2020few}
Vassilis~N Ioannidis, Da~Zheng, and George Karypis.
\newblock Few-shot link prediction via graph neural networks for covid-19 drug-repurposing.
\newblock \emph{arXiv preprint arXiv:2007.10261}, 2020{\natexlab{b}}.

\bibitem[Islam et~al.(2023)Islam, Amaya-Ramirez, Maigret, Devignes, Aridhi, and Sma{\"\i}l-Tabbone]{islam2023molecular}
Md~Kamrul Islam, Diego Amaya-Ramirez, Bernard Maigret, Marie-Dominique Devignes, Sabeur Aridhi, and Malika Sma{\"\i}l-Tabbone.
\newblock Molecular-evaluated and explainable drug repurposing for covid-19 using ensemble knowledge graph embedding.
\newblock \emph{Scientific Reports}, 13\penalty0 (1):\penalty0 3643, 2023.

\bibitem[Wishart et~al.(2008)Wishart, Knox, Guo, Cheng, Shrivastava, Tzur, Gautam, and Hassanali]{wishart2008drugbank}
David~S Wishart, Craig Knox, An~Chi Guo, Dean Cheng, Savita Shrivastava, Dan Tzur, Bijaya Gautam, and Murtaza Hassanali.
\newblock Drugbank: a knowledgebase for drugs, drug actions and drug targets.
\newblock \emph{Nucleic acids research}, 36\penalty0 (suppl\_1):\penalty0 D901--D906, 2008.

\bibitem[Himmelstein et~al.(2017)Himmelstein, Lizee, Hessler, Brueggeman, Chen, Hadley, Green, Khankhanian, and Baranzini]{himmelstein2017systematic}
Daniel~Scott Himmelstein, Antoine Lizee, Christine Hessler, Leo Brueggeman, Sabrina~L Chen, Dexter Hadley, Ari Green, Pouya Khankhanian, and Sergio~E Baranzini.
\newblock Systematic integration of biomedical knowledge prioritizes drugs for repurposing.
\newblock \emph{elife}, 6:\penalty0 e26726, 2017.

\bibitem[Percha and Altman(2018)]{percha2018global}
Bethany Percha and Russ~B Altman.
\newblock A global network of biomedical relationships derived from text.
\newblock \emph{Bioinformatics}, 34\penalty0 (15):\penalty0 2614--2624, 2018.

\bibitem[Del~Toro et~al.(2022)Del~Toro, Shrivastava, Ragueneau, Meldal, Combe, Barrera, Perfetto, How, Ratan, Shirodkar, et~al.]{del2022intact}
Noemi Del~Toro, Anjali Shrivastava, Eliot Ragueneau, Birgit Meldal, Colin Combe, Elisabet Barrera, Livia Perfetto, Karyn How, Prashansa Ratan, Gautam Shirodkar, et~al.
\newblock The intact database: efficient access to fine-grained molecular interaction data.
\newblock \emph{Nucleic acids research}, 50\penalty0 (D1):\penalty0 D648--D653, 2022.

\bibitem[Griffith et~al.(2013)Griffith, Griffith, Coffman, Weible, McMichael, Spies, Koval, Das, Callaway, Eldred, et~al.]{griffith2013dgidb}
Malachi Griffith, Obi~L Griffith, Adam~C Coffman, James~V Weible, Josh~F McMichael, Nicholas~C Spies, James Koval, Indraniel Das, Matthew~B Callaway, James~M Eldred, et~al.
\newblock Dgidb: mining the druggable genome.
\newblock \emph{Nature methods}, 10\penalty0 (12):\penalty0 1209--1210, 2013.

\bibitem[Cannon et~al.(2024)Cannon, Stevenson, Stahl, Basu, Coffman, Kiwala, McMichael, Kuzma, Morrissey, Cotto, et~al.]{cannon2024dgidb}
Matthew Cannon, James Stevenson, Kathryn Stahl, Rohit Basu, Adam Coffman, Susanna Kiwala, Joshua~F McMichael, Kori Kuzma, Dorian Morrissey, Kelsy Cotto, et~al.
\newblock Dgidb 5.0: rebuilding the drug--gene interaction database for precision medicine and drug discovery platforms.
\newblock \emph{Nucleic acids research}, 52\penalty0 (D1):\penalty0 D1227--D1235, 2024.

\bibitem[Ashburner et~al.(2000)Ashburner, Ball, Blake, Botstein, Butler, Cherry, Davis, Dolinski, Dwight, Eppig, et~al.]{ashburner2000gene}
Michael Ashburner, Catherine~A Ball, Judith~A Blake, David Botstein, Heather Butler, J~Michael Cherry, Allan~P Davis, Kara Dolinski, Selina~S Dwight, Janan~T Eppig, et~al.
\newblock Gene ontology: tool for the unification of biology.
\newblock \emph{Nature genetics}, 25\penalty0 (1):\penalty0 25--29, 2000.

\bibitem[Chambers et~al.(2013)Chambers, Davies, Gaulton, Hersey, Velankar, Petryszak, Hastings, Bellis, McGlinchey, and Overington]{chambers2013unichem}
Jon Chambers, Mark Davies, Anna Gaulton, Anne Hersey, Sameer Velankar, Robert Petryszak, Janna Hastings, Louisa Bellis, Shaun McGlinchey, and John~P Overington.
\newblock Unichem: a unified chemical structure cross-referencing and identifier tracking system.
\newblock \emph{Journal of cheminformatics}, 5\penalty0 (1):\penalty0 3, 2013.

\bibitem[Gaulton et~al.(2012)Gaulton, Bellis, Bento, Chambers, Davies, Hersey, Light, McGlinchey, Michalovich, Al-Lazikani, et~al.]{gaulton2012chembl}
Anna Gaulton, Louisa~J Bellis, A~Patricia Bento, Jon Chambers, Mark Davies, Anne Hersey, Yvonne Light, Shaun McGlinchey, David Michalovich, Bissan Al-Lazikani, et~al.
\newblock Chembl: a large-scale bioactivity database for drug discovery.
\newblock \emph{Nucleic acids research}, 40\penalty0 (D1):\penalty0 D1100--D1107, 2012.

\bibitem[Zdrazil et~al.(2024)Zdrazil, Felix, Hunter, Manners, Blackshaw, Corbett, de~Veij, Ioannidis, Lopez, Mosquera, et~al.]{zdrazil2024chembl}
Barbara Zdrazil, Eloy Felix, Fiona Hunter, Emma~J Manners, James Blackshaw, Sybilla Corbett, Marleen de~Veij, Harris Ioannidis, David~Mendez Lopez, Juan~F Mosquera, et~al.
\newblock The chembl database in 2023: a drug discovery platform spanning multiple bioactivity data types and time periods.
\newblock \emph{Nucleic acids research}, 52\penalty0 (D1):\penalty0 D1180--D1192, 2024.

\bibitem[Degtyarenko et~al.(2007)Degtyarenko, De~Matos, Ennis, Hastings, Zbinden, McNaught, Alc{\'a}ntara, Darsow, Guedj, and Ashburner]{degtyarenko2007chebi}
Kirill Degtyarenko, Paula De~Matos, Marcus Ennis, Janna Hastings, Martin Zbinden, Alan McNaught, Rafael Alc{\'a}ntara, Michael Darsow, Micka{\"e}l Guedj, and Michael Ashburner.
\newblock Chebi: a database and ontology for chemical entities of biological interest.
\newblock \emph{Nucleic acids research}, 36\penalty0 (suppl\_1):\penalty0 D344--D350, 2007.

\bibitem[Hastings et~al.(2016)Hastings, Owen, Dekker, Ennis, Kale, Muthukrishnan, Turner, Swainston, Mendes, and Steinbeck]{hastings2016chebi}
Janna Hastings, Gareth Owen, Adriano Dekker, Marcus Ennis, Namrata Kale, Venkatesh Muthukrishnan, Steve Turner, Neil Swainston, Pedro Mendes, and Christoph Steinbeck.
\newblock Chebi in 2016: Improved services and an expanding collection of metabolites.
\newblock \emph{Nucleic acids research}, 44\penalty0 (D1):\penalty0 D1214--D1219, 2016.

\bibitem[Bolton et~al.(2008)Bolton, Wang, Thiessen, and Bryant]{bolton2008pubchem}
Evan~E Bolton, Yanli Wang, Paul~A Thiessen, and Stephen~H Bryant.
\newblock Pubchem: integrated platform of small molecules and biological activities.
\newblock In \emph{Annual reports in computational chemistry}, volume~4, pages 217--241. Elsevier, 2008.

\bibitem[Li et~al.(2010)Li, Cheng, Wang, and Bryant]{li2010pubchem}
Qingliang Li, Tiejun Cheng, Yanli Wang, and Stephen~H Bryant.
\newblock Pubchem as a public resource for drug discovery.
\newblock \emph{Drug discovery today}, 15\penalty0 (23-24):\penalty0 1052--1057, 2010.

\bibitem[Kim et~al.(2025)Kim, Chen, Cheng, Gindulyte, He, He, Li, Shoemaker, Thiessen, Yu, et~al.]{kim2025pubchem}
Sunghwan Kim, Jie Chen, Tiejun Cheng, Asta Gindulyte, Jia He, Siqian He, Qingliang Li, Benjamin~A Shoemaker, Paul~A Thiessen, Bo~Yu, et~al.
\newblock Pubchem 2025 update.
\newblock \emph{Nucleic Acids Research}, 53\penalty0 (D1):\penalty0 D1516--D1525, 2025.

\bibitem[Schriml et~al.(2019)Schriml, Mitraka, Munro, Tauber, Schor, Nickle, Felix, Jeng, Bearer, Lichenstein, et~al.]{schriml2019human}
Lynn~M Schriml, Elvira Mitraka, James Munro, Becky Tauber, Mike Schor, Lance Nickle, Victor Felix, Linda Jeng, Cynthia Bearer, Richard Lichenstein, et~al.
\newblock Human disease ontology 2018 update: classification, content and workflow expansion.
\newblock \emph{Nucleic acids research}, 47\penalty0 (D1):\penalty0 D955--D962, 2019.

\bibitem[Schriml et~al.(2022)Schriml, Munro, Schor, Olley, McCracken, Felix, Baron, Jackson, Bello, Bearer, et~al.]{schriml2022human}
Lynn~M Schriml, James~B Munro, Mike Schor, Dustin Olley, Carrie McCracken, Victor Felix, J~Allen Baron, Rebecca Jackson, Susan~M Bello, Cynthia Bearer, et~al.
\newblock The human disease ontology 2022 update.
\newblock \emph{Nucleic acids research}, 50\penalty0 (D1):\penalty0 D1255--D1261, 2022.

\bibitem[Vastrik et~al.(2007)Vastrik, D'Eustachio, Schmidt, Joshi-Tope, Gopinath, Croft, de~Bono, Gillespie, Jassal, Lewis, et~al.]{vastrik2007reactome}
Imre Vastrik, Peter D'Eustachio, Esther Schmidt, Geeta Joshi-Tope, Gopal Gopinath, David Croft, Bernard de~Bono, Marc Gillespie, Bijay Jassal, Suzanna Lewis, et~al.
\newblock Reactome: a knowledge base of biologic pathways and processes.
\newblock \emph{Genome biology}, 8:\penalty0 1--13, 2007.

\bibitem[Milacic et~al.(2024)Milacic, Beavers, Conley, Gong, Gillespie, Griss, Haw, Jassal, Matthews, May, et~al.]{milacic2024reactome}
Marija Milacic, Deidre Beavers, Patrick Conley, Chuqiao Gong, Marc Gillespie, Johannes Griss, Robin Haw, Bijay Jassal, Lisa Matthews, Bruce May, et~al.
\newblock The reactome pathway knowledgebase 2024.
\newblock \emph{Nucleic acids research}, 52\penalty0 (D1):\penalty0 D672--D678, 2024.

\bibitem[Tanaka et~al.(2025)Tanaka, Chen, Belloni, Gisladottir, Kefeli, Patterson, Srinivasan, Zietz, Sirdeshmukh, Berkowitz, et~al.]{tanaka2025onsides}
Yutaro Tanaka, Hsin~Yi Chen, Pietro Belloni, Undina Gisladottir, Jenna Kefeli, Jason Patterson, Apoorva Srinivasan, Michael Zietz, Gaurav Sirdeshmukh, Jacob Berkowitz, et~al.
\newblock Onsides database: Extracting adverse drug events from drug labels using natural language processing models.
\newblock \emph{Med}, 2025.

\bibitem[Weininger(1988)]{smiles}
David Weininger.
\newblock Smiles, a chemical language and information system. 1. introduction to methodology and encoding rules.
\newblock \emph{Journal of Chemical Information and Computer Sciences}, 28\penalty0 (1):\penalty0 31--36, 1988.

\bibitem[Morgan(1965)]{MorganFinger}
Harry~L. Morgan.
\newblock The generation of a unique machine description for chemical structures—a technique developed at chemical abstracts service.
\newblock \emph{Journal of Chemical Documentation}, 5\penalty0 (2):\penalty0 107--113, 1965.

\bibitem[Rogers and Hahn(2010)]{RogersFinger}
David Rogers and Mathew Hahn.
\newblock Extended-connectivity fingerprints.
\newblock \emph{Journal of Chemical Information and Modeling}, 50\penalty0 (5):\penalty0 742--754, 2010.

\bibitem[Capecchi et~al.(2020)Capecchi, Probst, and Reymond]{capecchi2020one}
Alice Capecchi, Daniel Probst, and Jean-Louis Reymond.
\newblock One molecular fingerprint to rule them all: drugs, biomolecules, and the metabolome.
\newblock \emph{Journal of cheminformatics}, 12:\penalty0 1--15, 2020.

\bibitem[Schlichtkrull et~al.(2018)Schlichtkrull, Kipf, Bloem, Van Den~Berg, Titov, and Welling]{schlichtkrull2018modeling}
Michael Schlichtkrull, Thomas~N Kipf, Peter Bloem, Rianne Van Den~Berg, Ivan Titov, and Max Welling.
\newblock Modeling relational data with graph convolutional networks.
\newblock In \emph{The semantic web: 15th international conference, ESWC 2018, Heraklion, Crete, Greece, June 3--7, 2018, proceedings 15}, pages 593--607. Springer, 2018.

\bibitem[Vashishth et~al.(2019)Vashishth, Sanyal, Nitin, and Talukdar]{vashishth2019composition}
Shikhar Vashishth, Soumya Sanyal, Vikram Nitin, and Partha Talukdar.
\newblock Composition-based multi-relational graph convolutional networks.
\newblock \emph{arXiv preprint arXiv:1911.03082}, 2019.

\end{thebibliography}

\clearpage

\appendix

\section{Details on \dataset structure}\label{ap:details}
Table~\ref{tab:edge_type_mapping} shows the mapping between the interaction types in \dataset and the original DRKG with the corresponding sources. Tables~\ref{tab:edge_type_to_source_with_totals} and \ref{tab:node_type_to_source_with_totals} illustrate details on the data sources for edge and nodes, respectively. 

\begin{table}[h!]
 \caption{Mapping of interaction types to their corresponding database sources. Each row corresponds to a unified relation type, while the columns indicate the presence or absence of that relation in the original source databases, with the corresponding interaction name if that interaction is present. For example, the edge types labeled “Activator” and “Agonist” from the DGIdb dataset are consolidated into a single standardized \texttt{Activator} relation type in our dataset. A dash (“--”) denotes that the corresponding interaction type is not present in the given data source (*DrugHumGen, **HumGenHumGen).}.
  \label{tab:edge_type_mapping}
  \centering
  \scriptsize
  \begin{tabularx}{\textwidth}{@{}l *{7}{>{\centering\arraybackslash}X}@{}}
    \toprule
    \textbf{New interaction}
      & \textbf{DrugBank}
      & \textbf{GNBR}
      & \textbf{Hetionet}
      & \textbf{STRING}
      & \textbf{IntAct}
      & \textbf{DGIdb}
      & \textbf{bioarx} \\
    \midrule

    \texttt{Activator}
      & --
      & A+
      & --
      & --
      & --
      & \makecell[tc]{Activator\\Agonist}
      & -- \\
    \addlinespace

    \texttt{Blocker}
      & --
      & A-
      & --
      & --
      & --
      & \makecell[tc]{Antagonist,\\Blocker,\\Channel\\ Blocker}
      & -- \\
    \addlinespace

    \texttt{CMP\_BIND}
      & Target
      & B
      & CbG
      & --
      & \makecell[tc]{Direct \\Interaction,\\Association,\\Physical \\Interaction}
      & Binder
      & DHG* \\
    \addlinespace

    \texttt{ENZYME}
      & Enzyme
      & Z
      & --
      & --
      & --
      & --
      & -- \\
    \addlinespace

    \texttt{EXPRESSION}
      & --
      & E
      & --
      & Expression
      & --
      & --
      & -- \\
    \addlinespace

    \texttt{Regulation}
      & --
      & Rg
      & Gr$>$G
      & --
      & --
      & --
      & -- \\
    \addlinespace

    \texttt{UPREGULATION}
      & --
      & E+
      & CuG
      & --
      & --
      & --
      & -- \\
    \addlinespace

    \texttt{DOWNREGULATION}
      & --
      & E-, N
      & CdG
      & --
      & --
      & Inhibitor
      & -- \\
    \addlinespace

    \texttt{GENE\_BIND}
      & --
      & B
      & GiG
      & Binding
      & \makecell[tc]{Physical \\Assoc.,\\Assoc.,\\Dir. Int.}
      & --
      & \makecell[tc]{HGHG**} \\
    \addlinespace

    \texttt{TREATMENT}
      & Treats
      & T
      & CtD
      & --
      & --
      & --
      & -- \\
    \addlinespace

    \texttt{J\_c}
      & --
      & J
      & --
      & --
      & --
      & --
      & -- \\
    \addlinespace

    \texttt{J\_g}
      & --
      & J
      & --
      & --
      & --
      & --
      & -- \\
    \addlinespace

    \texttt{gene\_OTHER\_cmp}
      & --
      & --
      & --
      & --
      & --
      & Other
      & -- \\
    \addlinespace

    \texttt{gene\_OTHER\_gene}
      & --
      & --
      & --
      & Other
      & --
      & --
      & -- \\
    \bottomrule
  \end{tabularx}
\end{table}

\begin{table}[h!]
  \caption{Edge type to source.}
  \label{tab:edge_type_to_source_with_totals}
  
  \centering
  \scriptsize
  \setlength{\tabcolsep}{4pt}
  \renewcommand{\arraystretch}{1.3}
  \begin{tabularx}{\textwidth}{
    @{}>{\raggedright\arraybackslash}p{3cm} 
    *{9}{>{\centering\arraybackslash}X}@{}
  }
    \toprule
    \textbf{Relation}
      & \textbf{DGIDb}
      & \textbf{DrugBank}
      & \textbf{GNBR}
      & \textbf{Hetionet}
      & \textbf{IntAct}
      & \textbf{OnSIDES}
      & \textbf{STRING}
      & \textbf{bioarx}
      & \textbf{Total} \\
    \midrule
    Anatomy\--Gene        & 0       & 0         & 0       & 726\,156  & 0       & 0       & 0       & 0       & 726\,156  \\
    Compound\--Compound   & 0       & 1\,161\,176 & 0     & 6\,441    & 0       & 0       & 0       & 0       & 1\,167\,617\\
    Compound\--Disease    & 0       & 4\,502    & 68\,455  & 609       & 0       & 0       & 0       & 0       & 73\,566   \\
    Compound\--Gene       & 22\,750  & 8\,671    & 34\,469  & 41\,751   & 1\,563   & 0       & 0       & 24\,248  & 133\,452  \\
    Compound\--SideEffect & 0       & 0         & 0       & 138\,017  & 0       & 284\,235 & 0       & 0       & 422\,252  \\
    Disease\--Anatomy     & 0       & 0         & 0       & 3\,602    & 0       & 0       & 0       & 0       & 3\,602    \\
    Disease\--Disease     & 0       & 0         & 0       & 543       & 0       & 0       & 0       & 0       & 543      \\
    Disease\--Gene        & 0       & 0         &  65\,679       & 27\,936   & 0       & 0       & 0       & 458     & 94\,073   \\
    Disease\--Symptom     & 0       & 0         & 0       & 3\,357    & 0       & 0       & 0       & 0       & 3\,357    \\
    Gene\--Gene           & 0       & 0         & 23\,994  & 461\,263  & 141\,926 & 0       & 723\,259 & 29\,523 & 1\,379\,965\\
    \midrule
    \textbf{Total}       & 22\,750  & 1\,174\,349 & 192\,597 & 1\,409\,675 & 143\,489 & 284\,235 & 723\,259 & 54\,229 & 4\,004\,583\\
    \bottomrule
  \end{tabularx}
\end{table}

\begin{table}[h!]

  \caption{Node type to source.}
  \label{tab:node_type_to_source_with_totals}

  \centering
  \scriptsize

  \begin{tabularx}{\textwidth}{
    @{}>{\raggedright\arraybackslash}p{3cm} 
    *{7}{>{\centering\arraybackslash}X}@{}
  }
    \toprule
    \textbf{Source}
      & \textbf{Anatomy}
      & \textbf{Compound}
      & \textbf{Disease}
      & \textbf{Gene}
      & \textbf{SideEffect}
      & \textbf{Symptom}
      & \textbf{Total} \\
    \midrule

    CHEBI                
      & 0       
      & 176     
      & 0       
      & 0       
      & 0       
      & 0       
      & 176     \\

    CHEMBL              
      & 0       
      & 77      
      & 0       
      & 0       
      & 0       
      & 0       
      & 77      \\

    DOID                
      & 0       
      & 0       
      & 2\,390  
      & 0       
      & 0       
      & 0       
      & 2\,390  \\

    MESH                
      & 0       
      & 0       
      & 2\,356  
      & 0       
      & 0       
      & 415     
      & 2\,771  \\

    NCBI                
      & 0       
      & 0       
      & 0       
      & 20\,844 
      & 0       
      & 0       
      & 20\,844 \\

    OMIM                
      & 0       
      & 0       
      & 31      
      & 0       
      & 0       
      & 0       
      & 31      \\

    PubChem\_Compounds  
      & 0       
      & 15\,302 
      & 0       
      & 0       
      & 0       
      & 0       
      & 15\,302 \\

    UBERON              
      & 400     
      & 0       
      & 0       
      & 0       
      & 0       
      & 0       
      & 400     \\

    bioarx              
      & 0       
      & 0       
      & 27      
      & 0       
      & 0       
      & 0       
      & 27      \\

    drugbank            
      & 0       
      & 0       
      & 0       
      & 62      
      & 0       
      & 0       
      & 62      \\

    molport             
      & 0       
      & 221     
      & 0       
      & 0       
      & 0       
      & 0       
      & 221     \\

    umls                
      & 0       
      & 0       
      & 0       
      & 0       
      & 5\,704  
      & 0       
      & 5\,704  \\

    zinc                
      & 0       
      & 53      
      & 0       
      & 0       
      & 0       
      & 0       
      & 53      \\

    \textbf{Total}      
      & 400     
      & 15\,829 
      & 4\,804  
      & 20\,906 
      & 5\,704  
      & 415     
      & 48\,058 \\
    \bottomrule
  \end{tabularx}

\end{table}

\section{Experimental Setup}\label{ap:setup}
To produce a fair comparison of our baselines, we made sure to find the optimal parameters for each combination of task and dataset. The experiments were carried out using 4 NVIDIA A40 GPUs with 48~GB of VRAM. The hypermarameter tuning to around one week of computation. The experiments shown in the main paper took around 48 hours of computation.

Our search grid included the following parameters:
\begin{itemize}
\item Linear layer size: {5, 10, 20, 30, 40}

\item Convolutional layer: {1, 2}

\item First convolutional layer size: {5, 10, 15, 20}

\item Second convolutional layer size: {5, 10}

\item Learning rate: {0.001, 0.01}

\item Regularization: {0.001, 0.01}

\item Number of bases: {4, 8, 16}

\item Epochs: {400}

\item Dropout: {0.1, 0.2, 0.3}

\item Operation type: {subtraction, multiplication, circular correlation}
\end{itemize}

The following tables from Table~\ref{tab:best_hyperparams_drkg1} through Table~\ref{tab:best_hyperparams_vitagraph3} list the parameters of the best models divided by task and the dataset used.

\begin{table}[ht]
  \caption{\small{Best hyperparameters side effect prediction in DRKG dataset}.}
  \label{tab:best_hyperparams_drkg1}
  \centering
  \scriptsize
  \begin{tabularx}{\textwidth}{
    @{}>{\raggedright\arraybackslash}p{1cm}
    *{9}{>{\centering\arraybackslash}X}@{}
  }
    \toprule
    \textbf{Model}
      & \textbf{\# Conv Layers}
      & \textbf{Conv\_1 Size}
      & \textbf{Conv\_2 Size}
      & \textbf{Linear Size}
      & \textbf{Learning Rate}
      & \textbf{\# Bases}
      & \textbf{Dropout}
      & \textbf{Op Type}
      & \textbf{Regularization} \\
    \midrule

    R-GCN
      & 2
      & 40
      & 15
      & 30
      & 0.01
      & 32
      & –
      & –
      & 0.001 \\

    R-GAT
      & 2
      & 15
      & 10
      & 40
      & 0.01
      & 4
      & –
      & –
      & 0.001 \\

    CompGCN
      & 1
      & 5
      & -
      & 40
      & 0.01
      & –
      & 0.3
      & mult
      & 0.01 \\

    \bottomrule
  \end{tabularx}
\end{table}

\begin{table}[ht]
  \caption{\small{Best hyperparameters for side effect prediction in \dataset with no features}.}
  \label{tab:best_hyperparams_no_features1}
  \centering
  \scriptsize
  \begin{tabularx}{\textwidth}{
    @{}>{\raggedright\arraybackslash}p{1cm}
    *{9}{>{\centering\arraybackslash}X}@{}
  }
    \toprule
    \textbf{Model}
      & \textbf{\# Conv Layers}
      & \textbf{Conv\_1 Size}
      & \textbf{Conv\_2 Size}
      & \textbf{Linear Size}
      & \textbf{Learning Rate}
      & \textbf{\# Bases}
      & \textbf{Dropout}
      & \textbf{Op Type}
      & \textbf{Regularization} \\
    \midrule

    R-GCN
      & 1
      & 10
      & -
      & 30
      & 0.01
      & 8
      & –
      & –
      & 0.001 \\

    R-GAT
      & 1
      & 5
      & -
      & 40
      & 0.01
      & 8
      & –
      & –
      & 0.001 \\

    CompGCN
      & 1
      & 15
      & -
      & 5
      & 0.001
      & –
      & 0.2
      & sub
      & 0.001 \\

    \bottomrule
  \end{tabularx}
\end{table}

\begin{table}[h!]
  \caption{\small{Best hyperparameters for side effect prediction in \dataset}.}
  \label{tab:best_hyperparams_vitagraph1}
  \centering
  \scriptsize
  \begin{tabularx}{\textwidth}{
    @{}>{\raggedright\arraybackslash}p{1cm}
    *{9}{>{\centering\arraybackslash}X}@{}
  }
    \toprule
    \textbf{Model}
      & \textbf{\# Conv Layers}
      & \textbf{Conv\_1 Size}
      & \textbf{Conv\_2 Size}
      & \textbf{Linear Size}
      & \textbf{Learning Rate}
      & \textbf{\# Bases}
      & \textbf{Dropout}
      & \textbf{Op Type}
      & \textbf{Regularization} \\
    \midrule

    R-GCN
      & 2
      & 10
      & 5
      & 40
      & 0.01
      & 16
      & –
      & –
      & 0.001 \\

    R-GAT
      & 1
      & 10
      & -
      & 5
      & 0.01
      & 4
      & –
      & –
      & 0.001 \\

    CompGCN
      & 1
      & 20
      & -
      & 40
      & 0.001
      & –
      & 0.2
      & sub
      & 0.01 \\

    \bottomrule
  \end{tabularx}
\end{table}

\begin{table}[h!]
  \caption{\small{Best hyperparameters for the prediction of PPI in DRKG}.}
  \label{tab:best_hyperparams_drkg2}
  \centering
  \scriptsize
  \begin{tabularx}{\textwidth}{
    @{}>{\raggedright\arraybackslash}p{1cm}
    *{9}{>{\centering\arraybackslash}X}@{}
  }
    \toprule
    \textbf{Model}
      & \textbf{\# Conv Layers}
      & \textbf{Conv\_1 Size}
      & \textbf{Conv\_2 Size}
      & \textbf{Linear Size}
      & \textbf{Learning Rate}
      & \textbf{\# Bases}
      & \textbf{Dropout}
      & \textbf{Op Type}
      & \textbf{Regularization} \\
    \midrule

    R-GCN
      & 1
      & 10
      & -
      & 50
      & 0.01
      & 16
      & –
      & –
      & 0.1 \\

    R-GAT
      & 1
      & 15
      & -
      & 10
      & 0.01
      & 4
      & –
      & –
      & 0.01 \\

    CompGCN
      & 1
      & 15
      & -
      & 30
      & 0.01
      & –
      & 0.1
      & mult
      & 0.01 \\

    \bottomrule
  \end{tabularx}
\end{table}

\begin{table}[h!]
  \caption{\small{Best hyperparameters for the prediction of PPI in \dataset with no features}.}
  \label{tab:best_hyperparams_no_features2}
  \centering
  \scriptsize
  \begin{tabularx}{\textwidth}{
    @{}>{\raggedright\arraybackslash}p{1cm}
    *{9}{>{\centering\arraybackslash}X}@{}
  }
    \toprule
    \textbf{Model}
      & \textbf{\# Conv Layers}
      & \textbf{Conv\_1 Size}
      & \textbf{Conv\_2 Size}
      & \textbf{Linear Size}
      & \textbf{Learning Rate}
      & \textbf{\# Bases}
      & \textbf{Dropout}
      & \textbf{Op Type}
      & \textbf{Regularization} \\
    \midrule

    R-GCN
      & 1
      & 15
      & -
      & 40
      & 0.01
      & 16
      & –
      & –
      & 0.01 \\

    R-GAT
      & 1
      & 15
      & -
      & 20
      & 0.01
      & 8
      & –
      & –
      & 0.001 \\

    CompGCN
      & 1
      & 5
      & -
      & 30
      & 0.01
      & –
      & 0.2
      & mult
      & 0.001 \\

    \bottomrule
  \end{tabularx}
\end{table}

\begin{table}[h!]
  \caption{\small{Best hyperparameters for the prediction of PPI in \dataset}.}
  \label{tab:best_hyperparams_vitagraph2}
  \centering
  \scriptsize
  \begin{tabularx}{\textwidth}{
    @{}>{\raggedright\arraybackslash}p{1cm}
    *{9}{>{\centering\arraybackslash}X}@{}
  }
    \toprule
    \textbf{Model}
      & \textbf{\# Conv Layers}
      & \textbf{Conv\_1 Size}
      & \textbf{Conv\_2 Size}
      & \textbf{Linear Size}
      & \textbf{Learning Rate}
      & \textbf{\# Bases}
      & \textbf{Dropout}
      & \textbf{Op Type}
      & \textbf{Regularization} \\
    \midrule

    R-GCN
      & 2
      & 10
      & 5
      & 10
      & 0.01
      & 4
      & –
      & –
      & 0.001 \\

    R-GAT
      & 1
      & 10
      & 
      & 40
      & 0.01
      & 16
      & –
      & –
      & 0.01 \\

    CompGCN
      & 1
      & 10
      & -
      & 10
      & 0.01
      & –
      & 0.1
      & mult
      & 0.001 \\

    \bottomrule
  \end{tabularx}
\end{table}

\begin{table}[h!]
  \caption{\small{Best hyperparameters for drug repurposing in DRKG}.}
  \label{tab:best_hyperparams_drkg3}
  \centering
  \scriptsize
  \begin{tabularx}{\textwidth}{
    @{}>{\raggedright\arraybackslash}p{1cm}
    *{9}{>{\centering\arraybackslash}X}@{}
  }
    \toprule
    \textbf{Model}
      & \textbf{\# Conv Layers}
      & \textbf{Conv\_1 Size}
      & \textbf{Conv\_2 Size}
      & \textbf{Linear Size}
      & \textbf{Learning Rate}
      & \textbf{\# Bases}
      & \textbf{Dropout}
      & \textbf{Op Type}
      & \textbf{Regularization} \\
    \midrule

    R-GCN
      & 2
      & 15
      & 5
      & 30
      & 0.01
      & 4
      & –
      & –
      & 0.01 \\

    R-GAT
      & 1
      & 10
      & -
      & 40
      & 0.01
      & 16
      & –
      & –
      & 0.01 \\

    CompGCN
      & 2
      & 15
      & 10
      & 10
      & 0.01
      & –
      & 0.2
      & mult
      & 0.001 \\

    \bottomrule
  \end{tabularx}
\end{table}

\begin{table}[h!]
  \caption{\small{Best hyperparameters for drug repurposing in \dataset with no features}.}
  \label{tab:best_hyperparams_no_features3}
  \centering
  \scriptsize
  \begin{tabularx}{\textwidth}{
    @{}>{\raggedright\arraybackslash}p{1cm}
    *{9}{>{\centering\arraybackslash}X}@{}
  }
    \toprule
    \textbf{Model}
      & \textbf{\# Conv Layers}
      & \textbf{Conv\_1 Size}
      & \textbf{Conv\_2 Size}
      & \textbf{Linear Size}
      & \textbf{Learning Rate}
      & \textbf{\# Bases}
      & \textbf{Dropout}
      & \textbf{Op Type}
      & \textbf{Regularization} \\
    \midrule

    R-GCN
      & 1
      & 10
      & -
      & 20
      & 0.01
      & 4
      & –
      & –
      & 0.001 \\

    R-GAT
      & 1
      & 10
      & -
      & 40
      & 0.01
      & 16
      & –
      & –
      & 0.01 \\

    CompGCN
      & 1
      & 5
      & -
      & 5
      & 0.01
      & –
      & 0.3
      & mult
      & 0.01 \\

    \bottomrule
  \end{tabularx}
\end{table}

\begin{table}[h!]
  \caption{\small{Best hyperparameters for drug repurposing in \dataset}.}
  \label{tab:best_hyperparams_vitagraph3}
  \centering
  \scriptsize
  \begin{tabularx}{\textwidth}{
    @{}>{\raggedright\arraybackslash}p{1cm}
    *{9}{>{\centering\arraybackslash}X}@{}
  }
    \toprule
    \textbf{Model}
      & \textbf{\# Conv Layers}
      & \textbf{Conv\_1 Size}
      & \textbf{Conv\_2 Size}
      & \textbf{Linear Size}
      & \textbf{Learning Rate}
      & \textbf{\# Bases}
      & \textbf{Dropout}
      & \textbf{Op Type}
      & \textbf{Regularization} \\
    \midrule

    R-GCN
      & 1
      & 5
      & -
      & 30
      & 0.001
      & 8
      & –
      & –
      & 0.01 \\

    R-GAT
      & 1
      & 20
      & 
      & 30
      & 0.01
      & 16
      & –
      & –
      & 0.01 \\

    CompGCN
      & 1
      & 15
      & -
      & 20
      & 0.01
      & –
      & 0.3
      & sub
      & 0.01 \\

    \bottomrule
  \end{tabularx}
\end{table}

\section{Data sources details}\label{ap:sources}
Our work is released under the Attribution-NonCommercial 4.0 International (CC BY-NC 4.0) license. However, due to data integration from diverse sources, the individual licenses of each data provider should be considered, applied on a per-node and per-edge basis, depending on the database of origin. We hereby list the sources used, along with their license:

\begin{itemize}
    \item Reactome: Data are licensed under the Creative Commons Public Domain Dedication (CC0). More information at \url{https://reactome.org/license}

    \item OnSIDES: The dataset is licensed under the MIT License. The license details can be found in the GitHub repository: \url{https://github.com/tatonetti-lab/onsides}

    \item Hetionet: Hetionet original data is licensed under the Creative Commons Public Domain Dedication (CC0). More information at \url{https://github.com/hetio/hetionet}. For the data it integrates from additional external resources, the respective licenses are provided at \url{https://github.com/dhimmel/integrate/blob/master/licenses/README.md}


    \item UniChem: Data are licensed under the Creative Commons Public Domain Dedication (CC0). More information at \url{https://jcheminf.biomedcentral.com/articles/10.1186/1758-2946-5-3}.

    \item DRKG: The project is released under the Apache-2.0 License. Being itself an integration of several data sources, details on the licenses of the data it uses are available at \url{https://github.com/gnn4dr/DRKG/blob/master/licenses/Readme.md}

\end{itemize}

\end{document}